\documentclass[letterpaper, 10 pt, conference]{ieeeconf}  % Comment this line out if you need a4paper

\IEEEoverridecommandlockouts                              % This command is only needed if 
\let\labelindent\relax
\usepackage{times}
\usepackage{epsfig}
\usepackage{graphicx}
\usepackage{amsmath}
\usepackage{amssymb}
\usepackage{multirow}
\usepackage{etoolbox}
\usepackage{enumitem}
\usepackage{color}
\usepackage{booktabs}
\usepackage{subcaption}
\usepackage{cancel}
\usepackage{comment}
\usepackage[noadjust]{cite}
\usepackage{booktabs, dcolumn}
\makeatletter
\renewcommand{\paragraph}{%
  \@startsection{paragraph}{4}%
  {\z@}{2.25ex \@plus 1ex \@minus .2ex}{-1em}%
  {\normalfont\normalsize\bfseries}%
}
\makeatother

\title{\LARGE \bf 
Real-time Semantic Segmentation with Fast Attention}

\author{
Ping Hu$^1$,~Federico Perazzi$^2$,~Fabian Caba Heilbron$^2$,~Oliver Wang$^2$,\\ ~Zhe Lin$^2$,~Kate Saenko$^1$,~Stan Sclaroff$^1$
\thanks{$^{1}$Ping Hu, Kate Saenko, and Stan Sclaroff are with Department of Computer Science, Boston University.}
\thanks{$^{1}$Fabian Caba Heilbron, Oliver Wang, Zhe Lin, and Federico Perazzi are with Adobe Research.}
}
\begin{document}

\maketitle
\thispagestyle{empty}
\pagestyle{empty}

%%%%%%%%%%%%%%%%%%%%%%%%%%%%%%%%%%%%%%%%%%%%%%%%%%%%%%%%%%%%%%%%%%%%%%%%%%%%%%%%
%%%%%%%%% BODY TEXT
\begin{abstract}
In deep CNN based models for semantic segmentation, high accuracy relies on rich spatial context (large receptive fields) and fine spatial details (high resolution), both of which incur high computational costs. 
In this paper, we propose a novel architecture that addresses both challenges and achieves state-of-the-art performance for semantic segmentation of high-resolution images and videos in real-time. 
The proposed architecture relies on our fast spatial attention, which is a simple yet efficient modification of the popular self-attention mechanism and captures the same rich spatial context at a small fraction of the computational cost, by changing the order of operations. 
Moreover, to efficiently process high-resolution input, we apply an additional spatial reduction to intermediate feature stages of the network with minimal loss in accuracy thanks to the use of the fast attention module to fuse features. 
We validate our method with a series of experiments, and show that results on multiple datasets demonstrate superior performance with better accuracy and speed compared to existing approaches for real-time semantic segmentation. 
On Cityscapes, our network achieves 74.4$\%$ mIoU at 72 FPS and 75.5$\%$ mIoU at 58 FPS  on a single Titan X GPU, which is~$\sim$50$\%$ faster than the state-of-the-art while retaining the same accuracy. 
\end{abstract}
\section{Introduction}

Semantic segmentation is a fundamental task in robotic sensing and computer vision, aiming to predict dense semantic labels for given images~\cite{mottaghi2014role,geiger2013vision,caesar2018coco,milan2018semantic,TangDPBS18,nekrasov2019real}. 
With the ability to extract scene contexts such as category, location, and shape of objects and stuff (everything else), semantic segmentation can be widely applied to many important applications like robots~\cite{kostavelis2015semantic,stenborg2018long,wada2019joint} and autonomous driving~\cite{cordts2016cityscapes,zhou2018automated,meyer2018deep}. 
For many of these applications, \emph{efficiency} is critical, especially in real-time ($\geq$30FPS) scenarios.
To achieve high accuracy semantic segmentation, previous methods rely on features enhanced with rich contextual cues~\cite{Zhao_2017_CVPR,chen2017deeplab,zhang2018context,fu2019dual,jiang2019dfnet} and high-resolution spatial details~\cite{zhao2018icnet,orsic2019defense}.
However, rich contextual cues are typically captured via very deep networks with sizable receptive fields~\cite{Zhao_2017_CVPR,chen2017deeplab,zhang2018context,fu2019dual} that require high computational costs; and detailed spatial information demand for inputs of high resolution~\cite{zhao2018icnet,orsic2019defense}, which incur high FLOPs during inference.

\iffalse
\begin{figure}[t!]
    \includegraphics[width=0.95
    \linewidth]{Figure/Intro.pdf}
    \caption{Speed and mean Intersection over Union (mIoU) comparison of real-time image semantic segmentation on Cityscapes. Included methods are ICNet~\cite{zhao2018icnet}, SwiftNet~\cite{orsic2019defense}, BiseNet~\cite{yu2018bisenet}, ShelfNet~\cite{Zhuang_2019_ICCV_Workshops}, ERFNet~\cite{romera2017efficient}, SegNet~\cite{badrinarayanan2017segnet}. Our FANet (denoted as red triangles) achieves state-of-the-art accuracy with much higher efficiency.}
    \label{fig:intro}
\end{figure}
\fi

Recent efforts have been made to accelerate models for real-time applications~\cite{mehta2018espnet,zhao2018icnet,orsic2019defense,yu2018bisenet,marin2019efficient,nekrasov2019real}. These efforts can be roughly grouped into two types. The first strategy is to adopt compact and shallow model architectures~\cite{zhao2018icnet,orsic2019defense,paszke2016enet,yu2018bisenet}.
However, this approach may decrease the model capacity and limit the size of the receptive field for features, therefore decreasing the model's discriminative ability. 
Another technique is to restrict the input to be low-resolution~\cite{paszke2016enet,yu2018bisenet,marin2019efficient}. 
Though greatly decreasing the computational complexity, low-resolution images may lose important details like object boundaries or small objects. 
As a result, both types of methods sacrifice effectiveness for speed, limiting their practical applicability.

In this work, we address these challenges by proposing the Fast Attention Network (FANet) for real-time semantic segmentation. 
To capture rich spatial contextual information, we introduce an efficient fast attention module. 
The original self-attention mechanism has been shown to be beneficial for various vision tasks~\cite{wang2018non,vaswani17att} due to its ability to capture non-local context from the input feature maps. 
However, given $c$ channels, the original self-attention~\cite{wang2018non,vaswani17att} has a computational complexity of $\mathcal{O}(n^2c)$, which is quadratic with respect to the feature's spatial size $n=height\times width$. 
In the task of semantic segmentation, where high-resolution feature maps are required, this is costly and limits the model's efficiency and applications to real-time scenarios. 
Instead, in our fast attention module, we replace the Softmax normalization used in self-attention with cosine similarity, thus converting the computation process to a series of matrix multiplication upon which the matrix-multiplication associativity can be applied to reduce the computational complexity to the linear $\mathcal{O}(nc^2)$, without loss of spatial information.
The proposed fast attention is $\frac{n}{c}$ times more efficient than the standard self-attention, given $n\gg c$ in semantic segmentation (e.g. $n$=128$\times$256 and c=512) . 

FANet works by first extracting different stages of feature maps, which are then enhanced by fast attention modules and finally merged from deep to shallow stages in a cascaded way for class label prediction.
Moreover, to process high-resolution inputs at real-time speed, we apply additional spatial reduction into FANet. Rather than directly down-scaling the input images, which loses spatial details, we opt for down-sampling intermediate feature maps. This strategy not only reduces computations but also enables lower layers to learn to extract features from high-resolution spatial details, enhancing FANet effectiveness.
% Moreover, to process high-resolution inputs at real-time speed, we apply additional spatial reduction into FANet. Rather than directly down-scaling the input images, which loses spatial details, we empirically show that down-sampling intermediate feature maps not only reduce computations, but also enables lower layers to learn to extract features from high-resolution spatial details, enhancing its effectiveness. 
As a result, with very low computational cost, FANet makes use of both rich contextual information and full-resolution spatial details. 
We conduct extensive experiments to validate our proposed approach, and the results on multiple datasets demonstrate that FANet can achieve the fastest speed with state-of-the-art accuracy when compared to previous approaches for real-time semantic segmentation. 
Furthermore, in pursuit of better performance for video streams, we generalize the fast attention module to spatial-temporal contexts, and show (in Sec. 4) that this has the same computational cost as the single-frame model and does not increase with the length of the temporal range.
This allows us to add rich spatial-temporal context to video semantic segmentation while avoiding an increase in computation. 

In summary, we contribute the following: (1)We introduce the fast attention module for non-local context aggregation for efficient semantic segmentation, and further generalize it to a spatial-temporal version for video semantic segmentation. (2) We empirically show that applying extra spatial reduction to intermediate feature stages of the network effectively decreases computational costs while enhancing the model rich spatial details. (3) We present a Fast Attention Network for real-time semantic segmentation of images and videos with state-of-the-art accuracy and much higher efficiency over previous approaches.

\section{Related Work}

%With the Fully Convolution Networks (FCNs)~\cite{} as pioneers for deep semantic segmentation, many following-up approaches have been proposed for pushing forward the state-of-the-art performance.
%~\cite{shuai2016dag,zheng2015conditional,Zhang_2017_Scale,Hung_2017_Scene,Yu_2018_CVPRLearning,Ke_2018_ECCV,He_2019_CVPR,Liu_2019_CVPR,Cheng_2019_SPGNetICCV,} 

Extracting rich context information is key for high-quality semantic segmentation~\cite{long2015fully,ding2018context,purkait2019seeing,chen2019fasterseg}. To this end, dilated convolutions~\cite{chen2016attention,yu2015multi} are proposed as an effective tool to enlarge receptive field without shrinking spatial resolution~\cite{li2017not,ding2018context}. DeepLab~\cite{chen2017deeplab} and PSPNet~\cite{Zhao_2017_CVPR} capture multi-scale spatial context. The encoder-decoder architecture is an effective way for extracting spatial context. Early works like  SegNet~\cite{badrinarayanan2017segnet} and U-net~\cite{ronneberger2015u} adopt the symmetric structures for encoder and decoder. RefineNet~\cite{Lin_2017_CVPR} designs the multi-path refinement module to enhance the feature maps from deep to shallow.  GCN~\cite{Peng_2017_CVPR,Zhang_2018_ECCV} explicitly refine predictions with large-kernel filters at different stages. Recently, Deeplab-v3+~\cite{chen2018encoder} integrates dilated convolution and spatial pyramid pooling into an encoder-decoder network to further boost the effectiveness. The self-attention~\cite{wang2018non,vaswani17att} mechanism has been applied in semantic segmentation~\cite{Li_2019_ICCV,He_2019_CVPR} with the superior ability to capture long-range dependency, which, however, may incur intensive computation. To achieve better efficiency, Zhu \textit{et al.}~\cite{Zhu_2019_ICCV} propose to sample sparse anchor pixel locations for saving computation. Huang  \textit{et al.}~\cite{Huang_2019_ICCV} only consider the pixels on the same column and row. Although these methods reduce computation, they all take an approximation of the self-attention model and only partially collect the spatial information. In contrast, our fast attention does not only greatly save computation, but also capture full information from the feature map without loss of spatial information. 

We also notice that there are several works on bilinear feature pooling~\cite{yue2018compact,chen20182} that are related to our fast attention. 
Yet, our work differentiates from them in three aspects. (1) ~\cite{yue2018compact,chen20182} approximate the affinity between pixels, while our fast attention is derived in a strictly equivalent form to built accurate affinity. (2) Unlike~\cite{yue2018compact,chen20182} that focus on recognition tasks, our fast attention effectively tackles the dense semantic segmentation task. (3) As we will show later, in contrast to ~\cite{yue2018compact,chen20182}, our fast attention allows for very efficient feature reuse in the video scenario, which can benefit video semantic segmentation with extra temporal context without increasing computation.

Existing methods for tackling video semantic segmentation can be grouped into two types. The first one~\cite{hu2020temporally,li2018low,jain2019accel,shelhamer2016clockwork,zhu17dff,Pohlen_2017_CVPR,xu2018dynamic,krevso2020efficient} takes advantage of the redundant information in video frames, and reduce computation by reusing the high-level feature computed at keyframes. These methods run very efficiently, while always struggling with the spatial misalignment between frames, which leads to a decreased accuracy. Differently, the other type of methods ignore the redundancy and focus on capturing the temporal context information from neighboring frames for better effectiveness~\cite{gadde2017semantic,jin2017video,nilsson2018semantic}, which, however, incurs extra computation to sharply decrease the efficiency. In contrast to these methods, our FANet can be easily extended to also aggregate temporal context and allow for efficient feature reuse, achieving both high effectiveness and efficiency.

\section{Fast Attention Network}

In this section, we describe the Fast Attention Network (FANet) for real-time image semantic segmentation. 
We start by presenting the fast attention module and analyzing its computational advantages over original self-attention. 
Then we introduce the architecture of FANet. 
Last, we show that extra spatial reduction at intermediate feature stages of the model enables us to extract rich spatial details from high-resolution inputs while keeping a low computational cost. 
%An overview of our FANet is shown in Fig.~\ref{fig:Arch} (a).

\begin{figure*}
\centering
\includegraphics[width=0.93\linewidth]{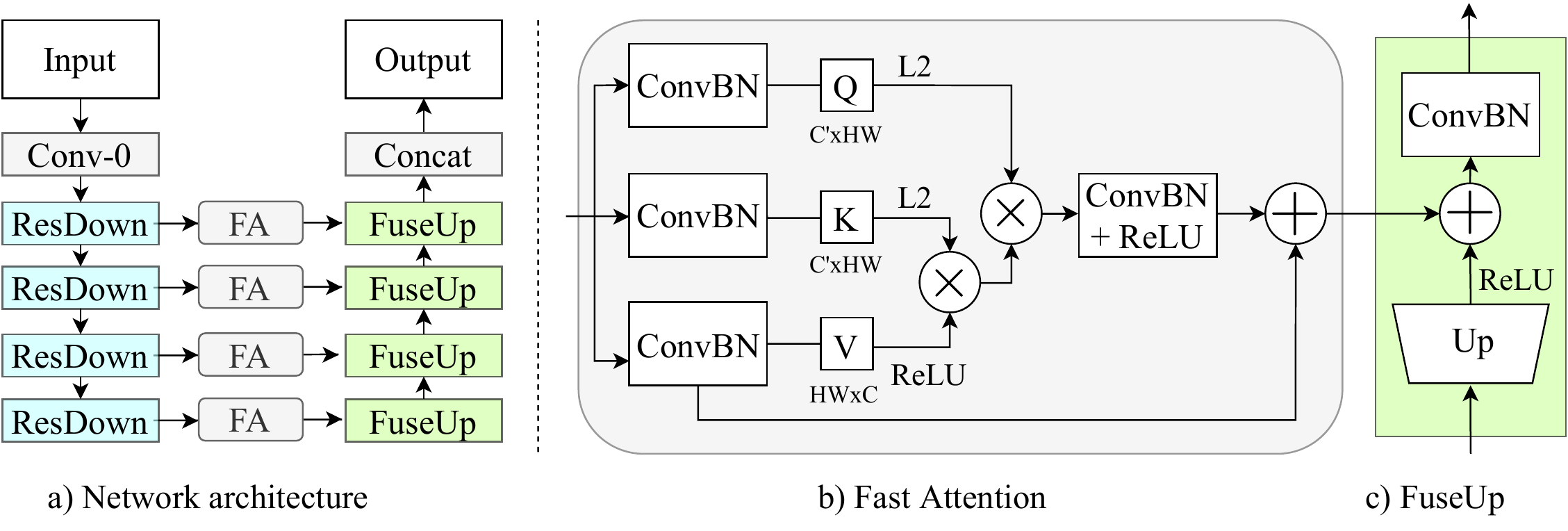}
\vspace{-0.3cm}
\caption{\small{(a) Architecture of Fast Attention Network (FANet). (b) Structure of Fast Attention (FA). (c) Structure of ``FuseUp'' module. Distinct from  channel  attention(CA) which only aggregates feature along the channel dimension for each  pixel independently, our Fast Attention aggregates  contextual information over the spatial domain  thus  achieving  better effectiveness }}
\label{fig:Arch} 
\vspace{-0.7cm}
\end{figure*}

\subsection{Fast Attention Module}

The self-attention module~\cite{wang2018non,vaswani17att} aims to capture non-local contextual information for each pixel location as a weighted sum of features at all positions in the feature map. 
Given a flattened input feature map $X\in\mathbb{R}^{n \times c}$ where $c$ is the channel size and $n=height\times width$ is the spatial size, the self-attention model~\cite{wang2018non,vaswani17att} applies 1$\times$1 convolutions to encode the feature maps into a \textit{\textbf{Value}} map $V\in\mathbb{R}^{n \times c}$ that contains the semantic information of each pixel, a \textit{\textbf{Query}} map $Q\in\mathbb{R}^{n \times c'}$ together with a \textit{\textbf{Key}} map $K\in\mathbb{R}^{n \times c'}$ that are used to build correlations between pixel positions.
Then the self-attention is calculated as $Y = f(Q,K)\cdot V$ where $f(\cdot,\cdot): \mathbb{R}^{n \times c'}\times\mathbb{R}^{n \times c'}\xrightarrow{}\mathbb{R}^{n \times n}$ is the \textbf{\textit{Affinity}} operation modeling the pairwise relations between all spatial locations.
The \textit{Softmax} function is typically used to model affinity $f(\cdot,\cdot)$, resulting in the popular self-attention response~\cite{Li_2019_ICCV,Zhu_2019_ICCV,Huang_2019_ICCV},
\begin{align}
    Y = Softmax(Q\cdot K^{\top})\cdot V
    \label{eq:softmax}
\end{align}

Due to the existence of the normalization term in the \textit{Softmax} function, the computation of Eq.~(\ref{eq:softmax}) needs first compute the inner matrix multiplication $Q\cdot K^{\top}$, and then the one outside.
This results in a computational complexity of $\mathcal{O}(n^2c)$.
In semantic segmentation, feature maps have high spatial resolution.
As complexity is quadratic with respect to the spatial size, this incurs high computational and memory costs, thus limiting the applications to scenarios especially those requiring real-time speed. 

%\begin{wrapfigure}{r}{.55\linewidth}
%\begin{figure}[t]
%\centering
%\includegraphics[clip,trim=0 0 0 0,width=0.7\linewidth]{./Figure/attention4.pdf}
%\caption{Standard self-attention (a), and our fast attention (b) modules.}
%\caption{Comparison between a) the self-attention mechanism proposed in~\cite{vaswani17att,wang2018non} and b) our Fast Attention module. Removing the Softmax normalization and swapping the order of matrix multiplications ((Q$\times$K)$\times$V vs. Q$\times$(K$\times$V)) captures the same contextual information at a fraction of the computational cost - up to 99$\%$ GFLOPs as shown in Table~\ref{tab:cmp_flops}.} 
%\end{figure}

We tackle this challenge by first removing the Softmax affinity. 
As indicated in~\cite{wang2018non}, there are a number of other affinity functions possible that can be used instead.
One example, the dot product affinity can be computed simply as: $f(Q,K)=Q\cdot K^\top$.
However, directly adopting the dot product may lead to affinity with unbounded values, and can be arbitrarily large. 
To avoid this, we instead use normalized cosine similarity for affinity computation,
\begin{align}
    Y = \frac{1}{n}(\hat{Q}\cdot\hat{K}^{\top})\cdot V
    \label{eq:cosine}
\end{align}
where $\hat{Q}$ and $\hat{K}$ are the results of $Q$ and $K$ after L2-normalization along the channel dimension. 
%For the attention in Eq.~\ref{eq:cosine} if we take the same computation process as in Eq.~\ref{eq:softmax}, the computational complexity is still  $\mathcal{O}(n^2c'+n^2c)$. 
Unlike Eq.~\ref{eq:softmax}, we observe that Eq.~\ref{eq:cosine} can be represented as a series of matrix multiplications, which means that we can apply standard matrix-multiplication associativity to change the order of computation to achieve our fast attention as follows, 
\begin{align}
    Y = \frac{1}{n}\hat{Q}\cdot(\hat{K}^{\top}\cdot V)
    \label{eq:assp}
\end{align}
where $n=height\times width$ is the spatial size, and $\hat{K}^{\top}\cdot V$ is computed first.

Without loss of generality, this fast attention module can be computed with a computational complexity of $\mathcal{O}(nc^2)$, which is only about $\frac{c}{n}$ of the computational requirement of Eq.~\ref{eq:softmax} (note that $n$ is typically much larger than $c$ in semantic segmentation). 
An illustration of fast attention module is shown in Fig.~\ref{fig:Arch} (b). We noticed that channel attention (CA)~\cite{fu2019dual} has similar computation to our FA, yet CA only aggregates feature along the channel dimension for each pixel, while our Fast Attention aggregates contextual information over the spatial  domain thus being more effective.

%FANet uses a ResNet18~\cite{he2016deep} encoder to generate the feature pyramid (left of a)). Feature maps at each stage is processed by a Fast Attention Module (FA) for efficient context aggregation (middle).  These feature maps are then merged to predict the final output in the decoder network (right).} \fch{Fix caption.}

\subsection{Network Architecture}
We describe our architecture for image semantic segmentation Fig~\ref{fig:Arch} (a). 
The network is an encoder-decoder architecture with three components: encoder (left), context aggregation (middle), and decoder (right). 
We use a light-weight backbone (ResNet18~\cite{he2016deep} without last fully connected layers) as the encoder to extract features from the input at different semantic levels.
Given an input with resolution $h\times w$, the first res-block (``Res-1'') in the encoder produces feature maps of $\frac{h}{4}\times\frac{w}{4}$ resolution. 
The other blocks sequentially output feature maps with resolution downsampled by a factor of 2.
Our network applies the fast attention modules at each stage. 
As shown in Fig.~\ref{fig:Arch} (b), the fast attention module is composed of three $1\times1$ convolutional layers for embedding the input features to be \textit{Query}, \textit{Key}, and \textit{Value} maps respectively. 
When generating the \textit{Query} and \textit{Key}, we remove the ReLU layer to allow for a wider range of correlation between pixels. 
The L2-normalization along the channel dimension makes sure the affinity is between -1 to +1. 
After the feature pyramid is processed by the fast attention modules, the decoder gradually merges and upsamples the features in a sequential fashion from deep feature maps to shallow ones. 
To enhance the decoded features with a high-level context, we further connect the middle features via a skip connection.  
An output with $\frac{h}{4}\times\frac{w}{4}$ resolution is predicted based on the enhanced feature output by the decoder.

%\begin{wrapfigure}{r}{0.55\textwidth}
%\centering
%\includegraphics[width=\linewidth]{./Figure/Downsample.pdf}
%\caption{\small{An example of applying extra spatial reduction to intermediate feature stages of FANet. Down-sampling intermediate feature stage not only saves computation for deeper layers but also enables lower layers to extract full-resolution spatial details}}
%\label{fig:Dsmpl} 
%\end{wrapfigure}

% \begin{figure}[t!]
% \centering
% \includegraphics[width=\linewidth]{./Figure/Dsmpl.pdf}
% \caption{\small{Examples of applying extra spatial reduction to FANet. ``Extra SR'' represents an extra spatial reduction. ``Extra US'' indicates an extra upsampling operation based on linear interpolation. Without additional down-sampling operation, (a) can access rich spatial information in the input, but has high computational cost. (b) Down-scaling the input by down-sampling or linear-interpolation helps save computation, but loses spatial details. (c) Down-sampling intermediate layers not only saves computation for deeper layers but also enables lower layers to extract full-resolution spatial details.} }
% \label{fig:Dsmpl} 
% \end{figure}
 
%\ow{shouldn't (b) have extra $\downarrow$ and $\uparrow$ at the top? I think this figure has very little information, we could just keep (c) and have it as a half width wrapfig} \fch{Agree with Oliver, I would only keep (c). Also make the figure consistent with the text: Extra DS $-->$ Extra SR (Spatial Reduction)}

\subsection{Extra Spatial Reduction for Real-time Speed}
\label{cnt:dsmpl}
Being able to generate semantic segmentation for high resolution inputs efficiently is challenging.
Typically, high-resolution inputs provide rich spatial details that help achieve better accuracy, but dramatically reduce efficiency~\cite{Zhao_2017_CVPR,chen2017deeplab,zhang2018context,Peng_2017_CVPR,chen2018encoder}.
On the other hand, using smaller input resolution saves computational costs, but generates worse results due to the loss of spatial details~\cite{paszke2016enet,yu2018bisenet,marin2019efficient}.

To alleviate this, we adopt a simple yet effective strategy, which is to apply additional down-sample operations to the intermediate feature stages of the network rather than directly down-sampling the input images.
We conduct an additional experiment where we use different types of spatial reduction operations, such as pooling and strided convolution at different feature stages, and evaluate how this impacts the resulting quality and speed trade-off. 
When applying an extra spatial reduction operator to our model, a similar up-sampling operation is added to the same stage of the decoder to keep the output resolution.
We select the best choice of these, which we show in Section~\ref{sec:ablations}, not only reduces computation for upper layers, but also allows lower layers to learn to extract rich spatial details from high-resolution inputs and enhance performance. Thus allowing for both real-time efficiency and effectiveness with full-resolution input.

\subsection{Extending to Video Semantic Segmentation}

In many real-world applications of semantic segmentation, such as self-driving and robotics, video streams are the natural input for vision systems to understand the physical world.
Nevertheless, most existing approaches for semantic segmentation focus on processing static images, and pay less attention to video data.
In addition to spatial context from individual frames, video sequences also contain important temporal context derived from dynamics in the camera and scene. 
To take advantage of such temporal context for better accuracy, in this section we extend our fast attention module to spatial-temporal contexts, and show that it improves video semantic segmentation without increasing computational costs.

Given $\{Q_{T},K_{T},V_{T}\}$ extracted from the target frame $T$, and $\{Q_{T-i},K_{T-i},V_{T-i}\}$ with $i\in\{1,2,..,t-1\}$ from the previous $t-1$ frames respectively, the spatial-temporal context within such a $t$-frame window can be aggregated via the traditional self-attention~\cite{wang2018non} as,
\begin{align}
    Y_T=\sum_{i=0}^{t-1} f(Q_{T}, K_{T-i})\cdot V_{T-i}.
\end{align}
This has a computational complexity of $\mathcal{O}(tn^2c)$, $t$ times higher than the single-frame spatial attention in Eq.~\ref{eq:softmax}. 

\begin{figure}
\centering
\includegraphics[width=0.88\linewidth]{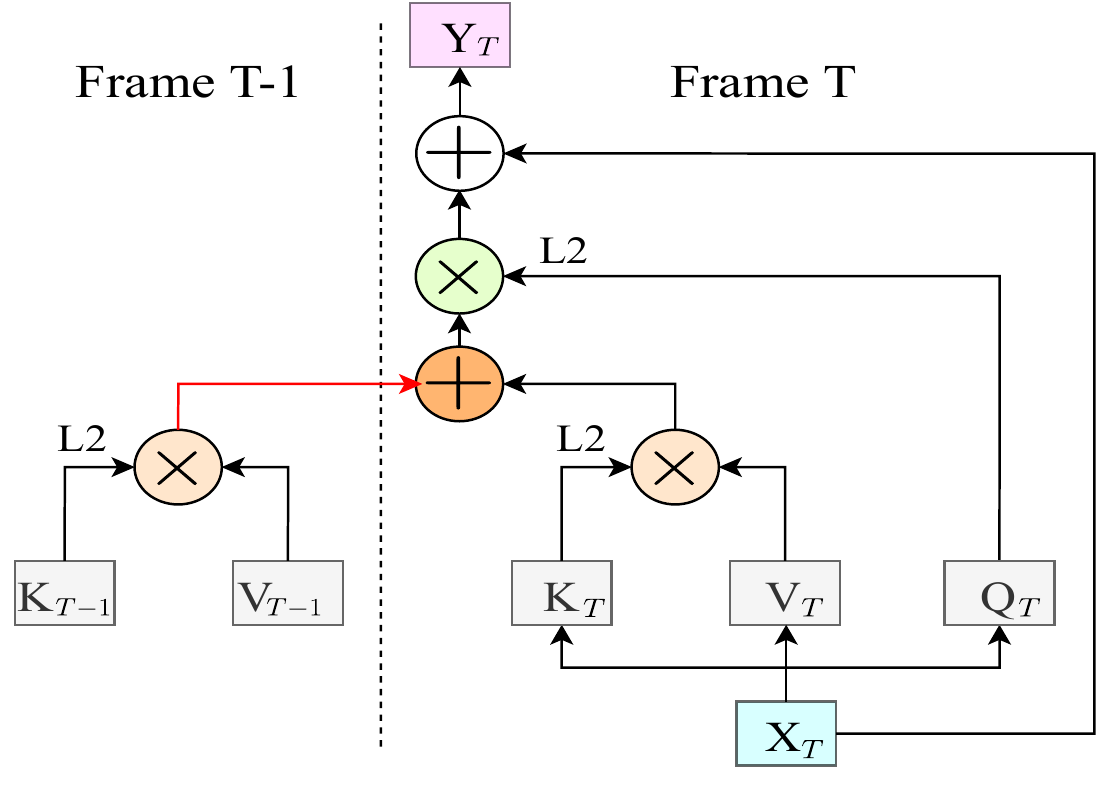}
\vspace{-0.4cm}
\caption{\small{Visualization of our fast attention for spatial-temporal context aggregation ($t$=2). The red arrows indicate the feature stored and reused by future frames.}}
\vspace{-0.3cm}
\label{fig:statn} 
\end{figure}

By replacing the original self-attention with our fast attention, the spatial-temporal context for the target frame $T$ can be computed as
\begin{align}
    Y_T&=\sum_{i=0}^{t-1} \frac{1}{n}\hat{Q}_{T}\cdot (\hat{K}_{T-i}^{\top}\cdot V_{T-i})\\
       %&=\frac{1}{n}\hat{Q}_{T}\cdot (\hat{K}_{T}^{\top}\cdot V_{T}) +\sum_{i=1}^{t-1} \frac{1}{n} \hat{Q}_{T}\cdot (\hat{K}_{T-i}^{\top}\cdot V_{T-i})\\
       &=\frac{1}{n}\hat{Q}_{T}\cdot (\hat{K}_{T}^{\top}\cdot V_{T}+\sum_{i=1}^{t-1} \hat{K}_{T-i}^{\top}\cdot V_{T-i})
    \label{eq:staatn}
\end{align}
where $n$ is the spatial size, $\hat{Q}$ and $\hat{K}$ indicate the L2-normalized $Q$ and $K$ respectively. 
At time step $T$, the results for $\hat{K}_{T-i}^{\top}\cdot V_{T-i}$ with $i\in\{1,2,...,t-1\}$ have already been computed and simply can be reused.
We can see in Eq.~\ref{eq:staatn}, that we only need to compute and store the term $\hat{K}_{T}^{\top}\cdot V_{T}$, add it to those of the previous frames' (this matrix addition's cost is negligible), and multiply it by $\hat{Q}_T$. 
Therefore, given a $t$-frame window our spatial-temporal fast attention has a computational complexity of $\mathcal{O}\left(n{c}^2\right)$, which is as efficient as the single-frame fast attention, and free of $t$. 
Therefore, our fast attention is able to aggregate spatial-temporal context without increasing computational cost.  An illustration of the spatial-temporal FA is shown in Fig.~\ref{fig:statn}. 
By replacing the fast attention modules with this spatial-temporal version, FANet is able to sequentially segment video frames with feature enhanced with spatial-temporal context.

%\fch{I would convert this section into a subsection of Sec 3. Perhaps, is there a way to write-up Eq6 in a generic form such it defaults to Eq4 when processing a single image?}
\section{Experiments}
%To demonstrate the effectiveness of the proposed FANet, we evaluate its performance on multiple benchmark datasets for scene understanding.
\subsection{Datasets and Evaluation}
\textit{Cityscapes}\cite{cordts2016cityscapes} is a large benchmark containing 19 semantic classes for urban scene understanding with 2975/500/1525 scenes for train/validation/test respectively. \textit{CamVid}\cite{brostow2008segmentation} is another street-view dataset with 11 classes. The annotated frames are divided into 367/101/233 for training/validation/testing. \textit{COCO-Stuff}\cite{caesar2018coco} contains both diverse indoor and outdoor scenes for semantic segmentation. This dataset has 9,000 densely annotated images for training and 1,000 for testing.  Following previous work\cite{zhao2018icnet}, we adopt the resolution 640$\times$640 and evaluate on 182 classes including 91 for things and 91 for stuff. 
We evaluate our method on image semantic segmentation for all four datasets, and additionally evaluate on Cityscapes for video semantic segmentation. The mIoU (mean Intersection over Union) is reported for evaluation.

%\myparagraph{ADE20K \cite{zhou2017scene}} is a challenging scene parsing benchmark with dense annotations for 150 object/stuff categories. 
%This dataset is split into 20K/2K/3K for training/validation/testing respectively. 
%On this dataset, we use 512$\times$512 resolution for both training and testing.\\

\subsection{Implementation Details}

We use ResNet-18/34~\cite{he2016deep} pretrained on Imagenet as the encoder in FANet, and randomly initialize parameters in fast attention modules as well as the decoder network.
We train using mini-batch stochastic gradient descent (SGD) with batchsize 16, weight decay 5e$^{-4}$, and momentum 0.9. The learning rate is initialized as 1e$^{-2}$, and multiplied with $(1-\frac{iter}{max\_iter})^{0.9}$ after each iteration.
We apply data augmentation including random horizontal flipping, random scaling (from 0.75 to 2), random cropping and color jittering in the training process.
During testing, we input images at full resolution, and resize the output to the original size for calculating the accuracy.
All the evaluation experiments are conducted with batchsize 1 on a single Titan X GPU. 
%\ow{can we mention anything about training time?}

%At first, we provide ablation study of AANet, then compare it with state-of-the-art approaches for real-time image semantic segmentation.

\begin{table}[]
\renewcommand\arraystretch{0.85}
\small
\begin{tabular}{p{1.7cm}p{0.6cm}p{0.6cm}p{0.6cm}p{0.6cm}p{0.6cm}p{0.6cm}} 
\toprule
\textit{C=}    &~~32   &~~64  &~128  &~256  &~512 &~1024\\
\midrule 
Self-Att.\cite{vaswani17att}  &~~68   &~103  &~173  &~313  &~602 &~1203\\
\textbf{Ours}         &~~0.2  &~~0.6 &~1.7   &~~5   &~~19 &~~~73\\
\bottomrule
\end{tabular}
\vspace{-0.1cm}
\caption{GFLOPs for non-local  module~\cite{wang2018non} and our fast attention module with C$\times$128$\times$256 features as input.}
\label{tab:cmp_flops}
\vspace{-0.6cm}
\end{table}

\iffalse
\begin{table}[t]
\renewcommand\arraystretch{0.85}
\small
\begin{tabular}{p{1.7cm}p{0.6cm}p{0.6cm}p{0.6cm}p{0.6cm}p{0.6cm}p{0.6cm}} 
\toprule
   \scriptsize{\textit{C=}}   &\scriptsize{~~32}        &\scriptsize{~~64} &\scriptsize{~128} &\scriptsize{~256} &\scriptsize{~512}&\scriptsize{~1024}\\
 \midrule 
\scriptsize{Self-Att.\cite{vaswani17att}}  &\scriptsize{~~35}     &\scriptsize{~~43}    &\scriptsize{~~64}  &\scriptsize{~~83}  &\scriptsize{~149}&\scriptsize{~~297}\\
\scriptsize{\textbf{Ours}}         &\scriptsize{~~0.8}    &\scriptsize{~~1.4}    &\scriptsize{~~~3}  &\scriptsize{~~~5}  &\scriptsize{~~12}&\scriptsize{~~~30}\\
\bottomrule
\end{tabular}
\vspace{2mm}
\caption{\footnotesize{Runtime (in ms) for non-local module~\cite{wang2018non} and our fast attention module with C$\times$128$\times$256 feature as input.}}
\label{tab:cmp_speed}
\vspace{-0.2cm}
\end{table}
\fi

\subsection{Method Analysis}
\label{sec:ablations}
\noindent\textbf{Fast Attention.} We first show the advantage in efficiency due to our fast attention. 
In Table~\ref{tab:cmp_flops}, we compare GFLOPs between a single original self-attention module and our fast attention module. 
Note that our fast attention runs significantly more efficiently for different size input features with more than 94$\%$ less computation.

We also compare our fast attention to the original self-attention module~\cite{vaswani17att} in our FANet. 
As shown in Table~\ref{tab:atn}, compared to the model without attention (denoted as ``w/o Att.''), applying the original self-attention module to the network increases mIoU by 2.4$\%$ while decreasing the speed from 83 fps to 8 fps.
In contrast to the original self-attention module, our fast attention (denoted as ``FA with L2-norm'') can achieve only slightly worse quality performance while greatly saving the computation cost.
To further analyze our cosine-similarity based fast attention, we also train without the L2-normalization for both \textit{Query} and \textit{Key} features (denoted as ``FA w/o L2-norm'') and achieve 74.1$\%$ mIoU on the Cityscapes $val$, which is lower than 75.0$\%$  mIoU of our full model. 
This validates the necessity of cosine similarity to ensure bounded values for affinity computation.

 \begin{table}[t!]
\centering
\small
\begin{tabular}{cccccc} 
\toprule
  $\#$Channel ($c$')    &~8  &~16  &~32  &~64  &~128\\ 
 \midrule 
 mIoU ($\%$) &73.5 &74.6 &75.0 &75.0 &75.0\\ 
 Speed (fps) &~74 &~74 &~72 &~69 &~65\\ 
\bottomrule
\end{tabular}
\vspace{-0.2cm}
\caption{\footnotesize{Performance on Cityscapes \textit{val} for different channel numbers ($c$') in fast attention in FANet-18.}}
\vspace{-0.4cm}
\label{tab:chn}
\end{table}

In Table~\ref{tab:chn}, we analyze the influence of channel numbers for \textit{Key} and \textit{Query} maps in our fast attention module. 
As we can see, too few channels such as $c$'=8 or $c$'=16 saves computation, but limits the representing capacity of the feature and leading to lower accuracy. On the other hand, when increasing the channel number from 32 to 128, the accuracy becomes stable, yet the speed drops. 
As a result, we adopt $c$'=32 in our experiments. 

\begin{table}
\centering
\begin{tabular}{cccc} 
\toprule
             &mIoU(\%)  &Speed(fps) &GFLOPs\\ 
 \midrule 
 w/o Att.    &~~72.7  &~~~83   &~~48\\ 
 Self-Att.\cite{vaswani17att}  &~~75.1  &~~~8    &~~121\\ 
 Channel-Att.\cite{fu2019dual} &~~74.6  &~~70    &~~51\\
 \midrule 
 FA w/o L2-norm &~~74.1  &~~~72   &~~49\\ 
 FA with L2-norm &~~75.0  &~~~72   &~~49\\ 
\bottomrule
\end{tabular}
\vspace{-0.2cm}
\caption{\small{Performance on Cityscapes for different attention mechanisms for FANet-18. ``FA'' denotes our fast attention.}}
\vspace{-0.8cm}
\label{tab:atn}
\end{table}

%\begin{wrapfigure}{r}{.5\linewidth}
%\centering
%\vspace{-.6cm}
%\includegraphics[clip,trim=0 16 400 0,width=\linewidth]{./Figure/DS_loc.pdf}
%\caption{\small{Results on Cityscapes (1024$\times$2048) with extra spatial reduction (rate=2) applied to different stages to FANet-18. Downsampling later in the encoder reduces running time, while improving result quality.}}
%\label{fig:DS_loc} 
%\end{wrapfigure}

\begin{figure}[h]
\centering
\includegraphics[width=\linewidth]{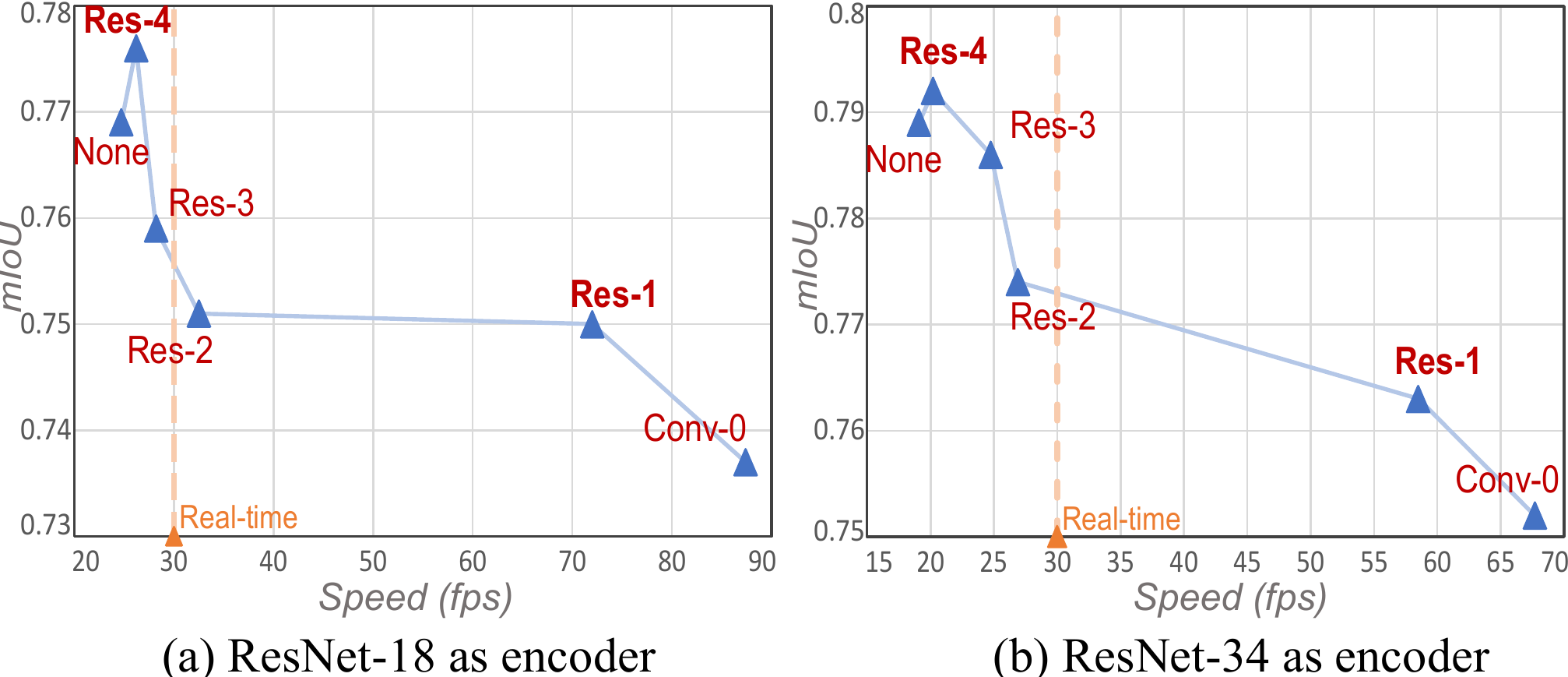}
\vspace{-0.6cm}
\caption{\small{Accuracy and speed analysis on Cityscapes \textit{val} for adding an additional down-sampling operation (rate=2) to different stages of the encoder in FANet. ``Conv-0'' means to directly down-sample the input image. ``Res-$n$'' indicates double the stride of the first Conv layer in the $n$-th Res-block. ``None'' means no additional down-sampling operation is applied.}}
\label{fig:DS_loc} 
\vspace{-0.4cm}
\end{figure}

\noindent\textbf{Spatial Reduction} Next, we analyze the effect of applying the extra spatial reduction at different feature stages of FANet. 
%We apply additional spatial down-sampling with rate 2 by doubling the stride of the first convolutional layer of the chosen network-block (Note that an linear interpolation based upsampling layer is applied at the same stage in decoder to keep the output resolution). 
The effects of additionally down-sampling different blocks are presented in Fig.~\ref{fig:DS_loc}. 
As we can see, down-sampling before ``Conv-0'' (down-scaling the input image), reduces the computation of all the subsequent layers, but loses critical spatial details which reduces the result quality. 
``Res-1'' indicates that we reduce the spatial size at the stage of the first Res-block in FANet. 
Extra spatial reduction at higher stages like ``Res-2'', ``Res-3'', and ``Res-4'' do not increase speed significantly.
Interestingly enough, we observe that applying down-sampling to ``Res-4'' actually performs \emph{better} than ``None'', no additional downsampling. 
We hypothesize that this may be because that the block ``Res-4'' process high-level features, and adding extra down-sampling helps to enlarge the receptive field thus benefiting with rich contextual information.
Based on these observations and with an aim of real-time semantic segmentation, we choose to apply extra down-sampling to ``Res-1'' and denote the model as FANet-18/34 based on the ResNet encoder used.
%\ow{why do we not use Res-4 for highest quality on Fig1? it looks like it would beat everything else? for speed/quality? or are there better slow networks?}
%, and apply extra down-sampling to ``Res-4'' for high-quality and denoted as AANet-slow.   \ow{why do we not do this?}
 
In additional to doubling the stride of convolutional layers to achieve 75.0$\%$ mIoU, we also experiment with other forms of down-sampling including average pooling (72.9$\%$ mIoU) and max pooling (74.2$\%$ mIoU). Enlarging stride for Conv layers performs the best.
This may be because that stride convolution helps to capture more spatial details while keeping sizable receptive fields.

\subsection{Image Semantic Segmentation}
We compare our final method to the recent state-of-the-art efficient approaches for real-time semantic segmentation. For fair comparisons, we evaluate the speed for different methods with PyTorch on the same Titan X GPU. 
Please check our supplementary material for details.
On benchmarks including Cityscapes~\cite{cordts2016cityscapes}, CamVid~\cite{brostow2008segmentation}, and COCO-Stuff~\cite{caesar2018coco}, our FANet achieves accuracy comparable to the state-of-the-art with the highest efficiency. 

%To fairly measure previous methods' speed, we test with models from the authors or our re-implementations. Check our  supplementary codes for details. 

\begin{table*}[!t]
\centering
%\scriptsize
\renewcommand\arraystretch{1}
\resizebox{.95\linewidth}{!}{
\begin{tabular}{p{2cm}p{0.9cm}p{0.9cm}p{1.8cm}p{1.4cm}p{2.5cm}p{1.6cm}}
\toprule
\multirow{2}{*}{~~~\footnotesize{Methods}}                                 & \multicolumn{2}{p{0.9cm}}{~~~~~~\footnotesize{mIoU($\%$)}} & \multirow{2}{*}{~~\footnotesize{Speed(fps)}} & \multirow{2}{*}{\footnotesize{GFLOPs}}& \multirow{2}{*}{GFLOPs\scriptsize{@1Mpx}} & ~~~\footnotesize{Input}  \\
\cline{2-3}
                                                            & ~~val                & ~~test          &                        &                    &                 & Resolution\\
\midrule 
~~\footnotesize{SegNet}~\cite{badrinarayanan2017segnet}     & ~~~~--               &~~56.1           &~~~~~~~36               &~~~~143             &~~~~~~{650}        &360$\times$640\\
%~~ESPNet-v2~\cite{mehta2019espnetv2}        & ~~~66.4   &~~~66.2      &~~~~\underline{57.5}     &~~~~~5.9     &512$\times$1024\\
~~\footnotesize{ICNet}~\cite{zhao2018icnet}                 & ~~67.7               &~~69.5           &~~~~~~~38               &~~~~~\textbf{30}    &~~~~~~~\textbf{15}   &1024$\times$2048\\
~~\footnotesize{ERFNet}~\cite{romera2017efficient}          & ~~71.5               &~~69.7           &~~~~~~~48               &~~~~103             &~~~~~~{206}         &512$\times$1024\\
~~\footnotesize{BiseNet}~\cite{yu2018bisenet}               & ~~74.8               &~~74.7           &~~~~~~~47               &~~~~~67             &~~~~~~{59.5}        &768$\times$1536\\
~~\footnotesize{ShelfNet}~\cite{Zhuang_2019_ICCV_Workshops} & ~~~~--               &~~\underline{74.8}&~~~~~~~39              &~~~~~95             &~~~~~~{47.5}        &1024$\times$2048\\
~~\footnotesize{SwiftNet}~\cite{orsic2019defense}           & ~~\underline{75.4}   &~~\textbf{75.5}   &~~~~~~~40              &~~~~106             &~~~~~~~{53}          &1024$\times$2048\\
\midrule 
~~\textbf{\footnotesize{FANet-34}}                          & ~~\textbf{76.3}      &~~\textbf{75.5}    &~~~~~~~\underline{58} &~~~~~65             &~~~~~~{32.5}  &1024$\times$2048\\
~~\textbf{\footnotesize{FANet-18}}                          & ~~75.0               &~~74.4             &~~~~~~~\textbf{72}    &~~~~~\underline{49} &~~~~~~\underline{24.5}  &1024$\times$2048\\
\bottomrule 
\end{tabular}}
\vspace{-0.2cm}
\caption{\small{ Image semantic segmentation performance comparison with recent state-of-the-art real-time methods on Cityscapes dataset. ``GFLOPs@1Mpx'' shows the GFLOPs for input with resolution 1M pixels.}}
\label{tab:cityscapes}
\vspace{-0.4cm}
\end{table*}

\begin{figure}
\centering
\includegraphics[width=0.96\linewidth]{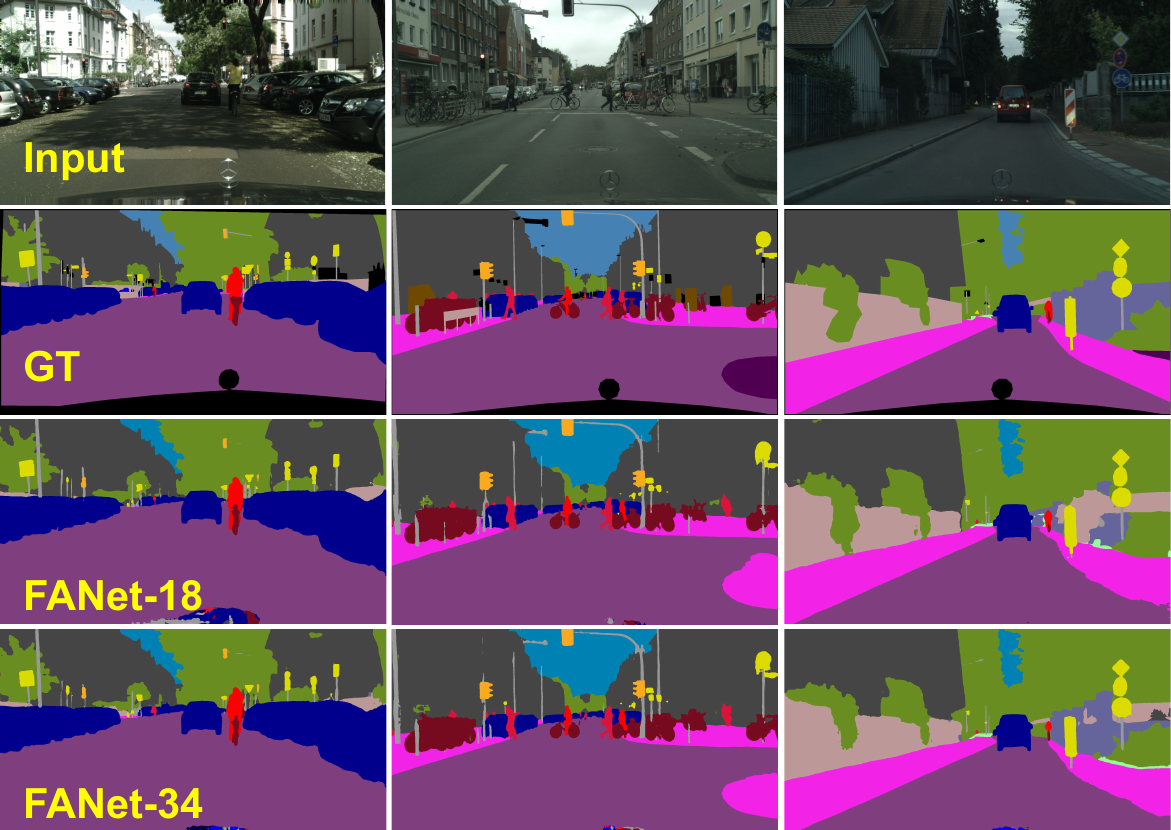}
\vspace{-0.2cm}
\caption{\small{Image semantic segmentation results on Cityscapes.}}
\label{fig:visual_res} 
\vspace{-0.2cm}
\end{figure}

\noindent\textbf{Cityscapes.} In Table~\ref{tab:cityscapes}, we present the speed-accuracy comparison. 
FANet-34 achieves mIoU 76.3$\%$ for validation and  75.5$\%$ for testing at a speed of 58 fps with full-resolution (1024$\times$2048) inputs. 
To our best knowledge, FANet-34 outperforms existing approaches for real-time semantic segmentation with better speed and state-of-the-art accuracy. 
By adopting a lighter-weight encoder ResNet-18, our FANet-18 further accelerates the speed to 72 fps, which is nearly two times faster than the recent methods like ShelfNet~\cite{Zhuang_2019_ICCV_Workshops} and SwiftNet~\cite{orsic2019defense}. 
Although the accuracy drops to mIoU 75.0$\%$ for validation and  74.4$\%$ for testing, it is still much better than many previous methods like SegNet~\cite{badrinarayanan2017segnet} and ICNet~\cite{zhao2018icnet}, and comparable to the most recent methods like BiseNet~\cite{yu2018bisenet} and ShelfNet~\cite{Zhuang_2019_ICCV_Workshops}. 
The performance achieved by our models demonstrates the superior ability to better balance the accuracy and speed for real-time semantic segmentation. Some visual results of our method are shown in Fig.~\ref{fig:visual_res}.

\noindent\textbf{CamVid.} Results for this dataset are reported in Table~\ref{tab:coco}. As we can see, our FANet outperforms previous methods with better accuracy and much faster speed. Comparing to BiseNet~\cite{yu2018bisenet}, our FANet-18 runs $2\times$ efficient, and our FANet-34 outperforms with $1.4\%$ mIoU and a faster speed.

\noindent\textbf{COCO-Stuff.} To be consistent with previous methods~\cite{zhao2018icnet}, we evaluate at resolution 640$\times$640 for segmenting the 182 categories. As shown in Table~\ref{tab:coco}, for the general scene understanding task with this dataset, our FANet is also able to achieve satisfying accuracy with much faster speed than previous methods. Compared to the state-of-the-art real-time model ICNet~\cite{zhao2018icnet}, our FANet-34 achieves both better accuracy and speed, and FANet-18 can further accelerate the speed with a comparable mIoU.

%\paragraph{ADE20K.}  In Table~\ref{tab:ade}, we report the validation set results. Our FANet performs with much higher efficiency than previous methods. Comparing to the state-of-the-art real-time method BiseNet~\cite{yu2018bisenet}, FANet-34 achieves better performance in both accuracy and speed, and  FANet-18 performs at a further faster speed over 200fps with comparable accuracy.

\begin{table}[!t]
\begin{minipage}[!t]{0.46\columnwidth}
\centering
\renewcommand\arraystretch{0.95}
\begin{tabular}{p{1.5cm}p{0.6cm}p{0.6cm}} 
\toprule
 \multirow{2}{*}{{Method}} & {mIoU} &{Speed}\\ 
   & {(\%)} &{(fps)}\\ 
 \midrule 
 {SegNet~\cite{badrinarayanan2017segnet}}    &~{55.6}              &{12}\\ 
 {ENet~\cite{paszke2016enet}}                &~{51.3}              &{46}\\ 
 {ICNet~\cite{zhao2018icnet}}                &~{67.1}              &{82}\\ 
 {BiseNet~\cite{yu2018bisenet}}              &~{68.7}              &{75}\\ 
 %\scriptsize{PSP50~\cite{Zhao_2017_CVPR}}               &~\underline{\scriptsize{69.1}}  &~\scriptsize{6}\\ 
 \midrule 
 {\textbf{FANet-34}}         &\textbf{~{70.1}}             &\underline{{121}}\\ 
 {\textbf{FANet-18}}         &~\underline{{69.0}}          &\textbf{{154}}\\ 
\bottomrule
\end{tabular}
\vspace{0mm}
\end{minipage}
\hspace{0.5cm}
\begin{minipage}[!t]{0.46\columnwidth}
\centering
\renewcommand\arraystretch{0.95}
\begin{tabular}{p{1.5cm}p{0.6cm}p{0.6cm}} 
\toprule
 \multirow{2}{*}{{Method}} & {mIoU} &{Speed}\\ 
   &{(\%)} &{(fps)}\\ 
 \midrule 
 {FCN~\cite{long2015fully}}               &~{22.7}              &{~~9}\\ 
 {DeepLab~\cite{chen2017deeplab}}         &~{26.9}              &{~14}\\ 
 {ICNet~\cite{zhao2018icnet}}             &~\underline{{29.1}}              &{110}\\
 {BiseNet~\cite{yu2018bisenet}}           &~{25.6}              &{113}\\ 
 %\scriptsize{PSP50~\cite{Zhao_2017_CVPR}}            &~\textbf{\scriptsize{32.6}}              &\scriptsize{~11}\\  
 \midrule 
 {\textbf{FANet-34}}                     &\textbf{~{29.5}}     &\underline{{142}}\\
 {\textbf{FANet-18}}                     &~{27.8}              &\textbf{{191}}\\ 
\bottomrule
\end{tabular}
\vspace{0mm}
\end{minipage}
\vspace{-0.1cm}
\caption{\small{Image semantic segmentation performance on Camvid (left) and COCO-Stuff (right).}}
\vspace{-0.7cm}
\label{tab:coco}
\end{table}

\begin{table}[h]
\renewcommand\arraystretch{1.1}
\setlength{\tabcolsep}{6pt}
\begin{tabular}{ccccc}
\toprule
{Method} & {mIoU} & {Speed} & {Avg RT} & {MaxLatency}\\ 
& ($\%$)$\uparrow$ & (fps)$\uparrow$ & (ms)$\downarrow$ & (ms)$\downarrow$ \\
 \midrule 
 {DVSNet-fast\cite{xu2018dynamic}}        &{63.2}              &{30.4}            &{33}              &- \\
 {Clockwork\cite{shelhamer2016clockwork}} &{64.4}              &{5.6}             &{177}             &{221} \\
 {DFF\cite{zhu17dff}}                     &{69.2}              &{5.7}             &{175}             &{644} \\
  {Accel\cite{jain2019accel}}  & 72.1 & 2.9 & 340 & 575 \\
 %{GRFP\cite{nilsson2018semantic}}         &{73.6}              &{3.5}             &{286}             &{286} \\
 {Low-Latency\cite{li2018low}}            &{75.9}              &{7.5}             &{133}             &{133}  \\
 %{PEARL\cite{jin2017video}}               &{76.5}  &{1.3}             &{800}             &{800}  \\
  {Netwarp\cite{gadde2017semantic}}& \textbf{80.6} &0.33  & 3004 & 3004\\
 \midrule 
 {FANet34}                                   &{76.3}              &\underline{58}             &\underline{17}             &\underline{17} \\
 {FANet34+Temp}                     &{\underline{76.7}}     &{\underline{58}} &{\underline{17}} &{\underline{17}} \\
 \midrule 
 {FANet18}                                   &{75.0}              &\textbf{72}             &\textbf{14}             &\textbf{14}\\
 {FANet18+Temp}                     &{75.5}              &{\textbf{72}}    &{\textbf{14}}    &{\textbf{14}}\\
\bottomrule
\end{tabular}
\vspace{-0.1cm}
\caption{\small{Video semantic segmentation on Cityscapes. ``+Temp'' indicates FANet with spatial-temporal attention (t=2). Avg RT is the average per-frame running time, and MaxLatency is the maximum per-frame running time.}}
\label{tab:vid}
\vspace{-0.6cm}
\end{table}

\subsection{Video Semantic Segmentation}
In this part, we evaluate our method for video semantic segmentation on the challenging dataset Cityscapes~\cite{cordts2016cityscapes}. Without significantly increasing the computational cost, our method can effectively capture both spatial and temporal contextual information to achieve better accuracy, and outperforms previous methods with much lower latency.  In Table~\ref{tab:vid}, we compare our method with recent state-of-the-art approaches for video semantic segmentation. 
Compared to the image segmentation baseline models FANet18 and FANet34, both our spatial-temporal version FANet18+Temp and FANet34+Temp help to improve the accuracy at the same computational costs.
We also see that most of the existing methods fail to achieve real-time speed ($\geq$ 30fps), apart from DVSNet which has much lower accuracy than ours.
Methods like Clockwork~\cite{shelhamer2016clockwork} and DFF~\cite{zhu17dff} save the overall computation while suffering from high latency due to the heavy computation at keyframes. PEARL~\cite{jin2017video} and Networp~\cite{gadde2017semantic} achieves state-of-the-art accuracy at the cost of very low speed and high latency. 
In contrast, FANet18+Temp and FANet34+Temp achieve state-of-the-art accuracy with a much faster speed. FANet18+Temp achieves more than 200$\times$ better efficiency than Netwarp~\cite{gadde2017semantic}. FANet34+Temp outperforms PEARL~\cite{jin2017video} with 40$\times$ faster speed.

%\paragraph{Cityscapes.} In Table~\ref{tab:vid}, we compare our method with recent state-of-the-art approaches for video semantic segmentation. 
%``+Temp'' indicates FANet with spatial-temporal associative attention for processing video frames. ``Avg RT'' represents the average per-frame time cost. ``Max Latency'' means the maximum per-frame time cost.
%Compared to the image segmentation baseline models FANet18 and FANet34, both our spatial-temporal version FANet18+Temp and FANet34+Temp help to improve the accuracy at the same computational costs.
%We also see that most of the existing methods fail to achieve real-time speed ($\geq$ 30fps), apart from DVSNet which has lower accuracy than ours. 
%Methods like Clockwork~\cite{shelhamer2016clockwork} and DFF~\cite{zhu17dff} save the overall computation while suffering from high latency due to the heavy computation at keyframes. PEARL~\cite{jin2017video} achieves state-of-the-art accuracy at the cost of very low speed and high latency. 
%In contrast, FANet18-Temp and FANet34-Temp achieve state-of-the-art accuracy with a much faster speed. 
%FANet34+Temp outperforms PEARL~\cite{jin2017video} with 40$\times$ faster speed.
 
%\paragraph{CamVid.} We also evaluate our method on CamVid dataset in Table~\ref{tab:vid}. 
%Again, spatial-temporal fast attention is effective at this task. 
%The consistent performance on both datasets shows that our method is robust to dataset domain. 

\section{Conclusion}
We have proposed a novel Fast Attention Network for real-time semantic segmentation. 
In the network, we introduce fast attention to efficiently capture contextual information from feature maps.
%The features of different stages are processed by fast attention modules and then merged to be the final output in a cascaded way. To ensure high-resolution input while keeping high efficiency, we also apply extra spatial reduction to the intermediate layers. 
We further extend the fast attention to spatial-temporal context, and apply our models to achieve low-latency video semantic segmentation.  To ensure high-resolution input with high efficiency, we also propose to apply spatial reduction to the intermediate feature stages. As a result, our model is enhanced with both rich contextual information and high-resolution details, while keeping a real-time speed.
Extensive experiments on multiple datasets demonstrate the efficiency and effectiveness of our method.

{\small
\bibliographystyle{IEEEtran}
\bibliography{IEEEabrv,egbib}
}

\end{document}